\documentclass[10pt,twocolumn,letterpaper]{article}

\usepackage{iccv}
\usepackage{times}
\usepackage{epsfig}
\usepackage{graphicx}
\usepackage{amsmath}
\usepackage{amssymb}
\usepackage{booktabs}
\usepackage{multirow}
\usepackage{pifont}
\usepackage{comment}

\usepackage[breaklinks=true,bookmarks=false]{hyperref}

\iccvfinalcopy 

\def\iccvPaperID{11} 

\makeatletter
\newcommand{\figcaption}[1]{\def\@captype{figure}\caption{#1}}
\newcommand{\tblcaption}[1]{\def\@captype{table}\caption{#1}}
\makeatother

\ificcvfinal\fi

\begin{document}

\title{BiLMa: Bidirectional Local-Matching for Text-based Person Re-identification}

\author{Takuro Fujii\\
Yokohama National University\\
{\tt\small tkr.fujii.ynu@gmail.com}
\and
Shuhei Tarashima\\
NTT Communications Corporation\\
{\tt\small tarashima@acm.org}
}

\maketitle

\begin{abstract}
Text-based person re-identification (TBPReID) aims to retrieve person images represented by a given textual query. 
In this task, how to effectively align images and texts globally and locally is a crucial challenge. 
Recent works have obtained high performances by solving Masked Language Modeling (MLM) to align image/text parts. 
However, they only performed uni-directional (i.e., from image to text) local-matching, leaving room for improvement by introducing opposite-directional (i.e., from text to image) local-matching.
In this work, we introduce Bidirectional Local-Matching (BiLMa) framework that jointly optimize MLM and Masked Image Modeling (MIM) in TBPReID model training.
With this framework, our model is trained so as the labels of randomly masked both image and text tokens are predicted by unmasked tokens. 
In addition, to narrow the semantic gap between image and text in MIM, we propose Semantic MIM (SemMIM), in which the labels of masked image tokens are automatically given by a state-of-the-art human parser.
Experimental results demonstrate that our BiLMa framework with SemMIM achieves state-of-the-art Rank@1 and mAP scores on three benchmarks.
\end{abstract}

\vspace{-3mm}
\section{Introduction}
Text-based person re-identification (TBPReID)~\cite{CUHK-PEDES} aims to retrieve a target person from an image pool given a textual query.
Since text queries are more user-friendly than image queries, TBPReID has been more and more expected to benefit various applications of surveillance and public safety.
\begin{figure}[t]
\centering
\includegraphics[width=8.3cm]{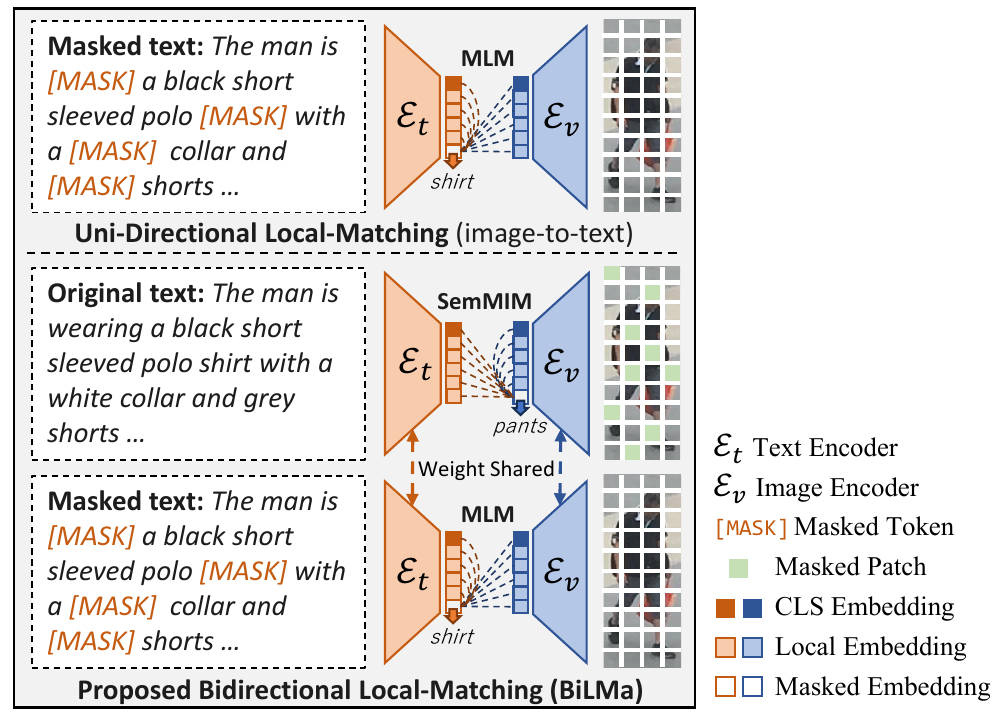}
   \caption{Overview of widely-used Uni-Directional Local-Matching and our Bidirectional Local-Matching (BiLMa). BiLMa exploits clues from both images and texts.}
\label{fig:overview}
\vspace{-5mm}
\end{figure}
Existing literatures focus on how to align images and texts globally~\cite{Zheng2017DualpathCI, Zhang_2018_ECCV} and/or locally~\cite{Li2021LearningSF,Chen2021TIPCBAS}.
Particularly, recent works have demonstrated the importance of image-text local-matching~\cite{10.1145/3503161.3548028,Yan2022CLIPDrivenFT}, and state-of-the-art (SOTA) methods~\cite{Jiang_2023_CVPR,Zuo2023PLIPLP,Bai2023RaSaRA} employ Masked Language Modeling (MLM) to align parts between image and text.
\par
Note that, in these MLM-based TBPReID methods, a model is trained via predicting the labels of masked text tokens using unmasked image and text tokens as shown in the top of Figure \ref{fig:overview}.
We argue that, however, these methods does not fully exploit local alignment between images and texts, because matching is performed only {\it uni-directionally} ({\it i.e.}, from image to text).
Local-matching of the opposite direction ({\it i.e.}, from text to image) could also contribute to align semantically similar local image tokens ({\it i.e.}, patches) with corresponding text parts, but this research direction has not been explored in the literature.
\par
In this work, we propose Bidirectional Local-Matching (BiLMa) framework that can enhance local image-text alignment by jointly optimizing image-to-text MLM and text-to-image Masked Image Modeling (MIM), as illustrated in the bottom of Figure \ref{fig:overview}.
In our BiLMa framework, TBPReID models are trained through predicting the labels of randomly masked both image and text tokens by all the unmasked tokens.
\par
Notice that a straightforward approach to perform MIM in our BiLMa is to adapt existing methods~\cite{9880205,Bao2021BEiTBP,Cao2022HowTU,Wang_2023_CVPR}, which are formulated as reconstruction problems.
However, we empirically found that solving reconstruction in TBPReID training is difficult ({\it cf.} \S A.5 in the supplementary material), since it suffers huge semantic gap between modalities.
To address this issue, we additionally propose a novel MIM method, named Semantic MIM (SemMIM).
In our SemMIM, we formulate the MIM as a prediction of semantic labels for randomly masked image tokens by unmasked image and text tokens.
With a SOTA human parser~\cite{li2020self}, we show that the semantic labels of tokens ({\it i.e.}, patches) can be automatically obtained.
\par
Experimental results demonstrate that our BiLMa with SemMIM achieves SOTA Rank@1 and mAP scores on three TBPReID benchmarks.
We also show that incorporating both MLM and MIM in TBPReID training ({\it i.e.}, BiLMa framework) leads to higher performances than the models with either MLM or MIM.
To summarize, our contributions are threefold:
(1) We propose Bidirectional Local-Matching (BiMLa) framework that jointly optimize MLM and MIM in TBPReID training.
(2) We propose Semantic MIM (SemMIM) that can make MIM in TBPReID training tractable.
(3) Experimental results demonstrate that our BiLMa with SemMIM achieves SOTA Rank@1 and mAP on three public benchmarks.

\section{Related Work}
\noindent\textbf{Text-based Person Re-identification (TBPReID).} 
This task was firstly introduced by \cite{CUHK-PEDES} with a benchmark dataset.
In this line of research, various solutions~\cite{Hoffer2014DeepML,Zhang_2018_ECCV,Li2021LearningSF,Yan2022CLIPDrivenFT} have been proposed, sparked by progress in the Vision-and-Language field.
Particularly, recent works have achieved state-of-the-art performances by introducing Masked Language Modeling (MLM). 
PLIP~\cite{Zuo2023PLIPLP} predicts masked textual tokens by masked textual tokens and visual tokens to construct the correlation between images and texts.
IRRA~\cite{Jiang_2023_CVPR} predicts masked tokens by the rest of unmasked textual tokens and visual tokens to align image and text contextualized representations and to model local dependencies.
However, their MLM methods perform only uni-directional local-matching ({\it i.e.}, from image to text), leaving room for improvement by introducing opposite-directional ({\it i.e.}, from text to image) local-matching.
In our work, we implement bidirectional local-matching to locally align images and texts more strongly by jointly optimizing image-to-text MLM and text-to-image MIM. 


\noindent\textbf{Masked Image Modeling (MIM).} 
MIM is originally designed for self-supervised visual learning.
There are various MIM strategies~\cite{Cao2022HowTU,9880205,Bao2021BEiTBP}, all of which are formulated as reconstruction problems of randomly masked visual tokens ({\it i.e.}, patches) by unmasked tokens.
However, we empirically found that image reconstruction from texts is difficult and not effective in TBPReID due to huge semantic gap between modalities ({\it cf.} \S A.5).
In our work, we design a novel MIM strategy, named Semantic MIM (SemMIM), for TBPReID to predict semantic labels of masked patches with textual and visual tokens.






\section{Method}
\begin{figure}[htbp]
\centering
\includegraphics[width=7cm]{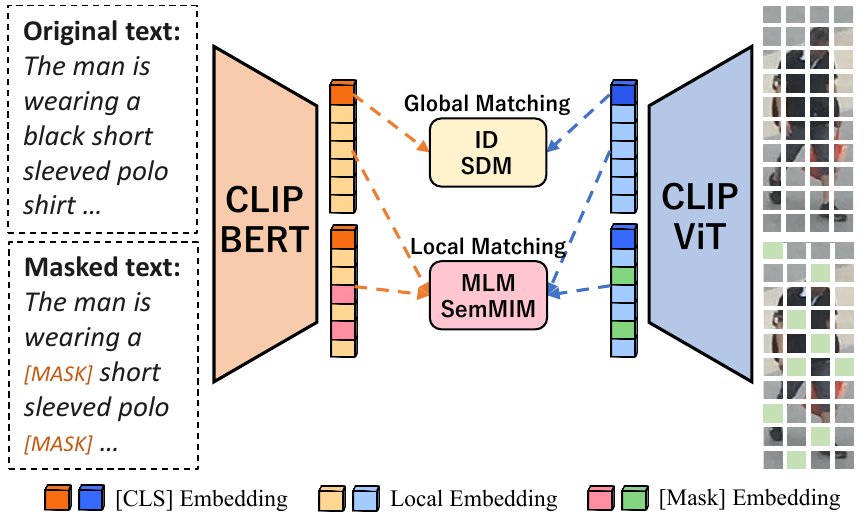}
   \caption{Overview of our BiLMa that uses ID and SDM loss for global-matching and MLM and SemMIM for local-matching.}
\label{fig:model}
\vspace{-4mm}
\end{figure}
Our BiLMa framework can be easily deployed on top of any Transformer-based vision-language models.
As a proof-of-concept, here we build BiLMa models based on IRRA~\cite{Jiang_2023_CVPR}, which is a SOTA TBPReID model at the time of this submission. 
In this section, we first introduce IRRA briefly, then detail our proposed BiLMa and SemMIM.

\subsection{IRRA~\cite{Jiang_2023_CVPR}}
IRRA is based on CLIP \cite{Radford2021LearningTV} image/text encoders.
The image encoder takes an image $I$ to produce a sequence of visual tokens, each of which represents a non-overlapping local token or a learnable \texttt{[CLS]} embedding.
We represent the output of image encoder as $\boldsymbol{h}^V=\{\boldsymbol{h}^V_{cls},\boldsymbol{h}^V_1,...,\boldsymbol{h}^V_{N_v}\}$, where $N_v$ is the number of tokens.
Similarly, the text encoder takes an input text to produce a sequence of text tokens, each of which corresponds to a subword token or \texttt{[SOS]}/\texttt{[EOS]} tokens.
The output of the text encoder is represented as $\boldsymbol{h}^T=\{\boldsymbol{h}^T_{sos},\boldsymbol{h}^T_1,...,\boldsymbol{h}^T_{N_t},\boldsymbol{h}^T_{eos}\}$, where $N_t$ is its token length.
\par
IRRA employs Masked Language Modeling (MLM) to train the whole model.
Specifically, during training, IRRA randomly replaces a portion of text tokens as a learnable \texttt{[MASK]} tokens.
All the unmasked $\boldsymbol{h}^V$ and $\boldsymbol{h}^T$ tokens are fed into an extra encoder, which produces embeddings for correctly predicting the labels of masked tokens. 
\par
The loss function to train IRRA models is the weighted sum of SDM loss $\mathcal{L}_{sdm}$~\cite{Jiang_2023_CVPR} 
and ID loss $\mathcal{L}_{id}$~\cite{Zheng2017DualpathCI} for global-matching, and MLM loss $\mathcal{L}_{mlm}$ for local-matching.
SDM loss is a KL-divergence between cosine similarity distributions of image-text pairs in mini-batch and true distribution, and
ID loss is instance-level intra-modal matching loss.
Please refer to Equation (1)-(4) in our supplementary material and the original papers \cite{Jiang_2023_CVPR,Zheng2017DualpathCI} for more details.
The MLM loss is a sum of Cross-Entropy between masked textual tokens and its labels, which is defined as follows:
\vspace{-1mm}
\begin{equation}
\small
    \mathcal{L}_{mlm}=-\frac{1}{|\mathcal{M}_t||\mathcal{V}|}\sum_{i\in \mathcal{M}_t}\sum_{j\in |\mathcal{V}|}y_j^i\log\frac{\exp(m_{i,j}^{T_m})}{\sum_{k=1}^{|\mathcal{V}|}\exp(m_{i,k}^{T_m})},
    \vspace{-1mm}
\end{equation}
where $\mathcal{M}_t$ denotes the set of masked textual tokens and $\mathcal{V}$ is the text vocabulary.
$y_j^i$ is 1 if the true label of $i$-th masked token is $j$-th vocablary in $\mathcal{V}$, and 0 otherwise.
$\{m_{i,j}^{T_m}\}_{j=1}^{|\mathcal{V}|}$ is the probability of $j$-th word in $\mathcal{V}$ of $i$-th masked textual token. 
\subsection{Bidirectional Local-Matching (BiLMa)}
\begin{figure}[htbp]
\centering
\includegraphics[width=8.3cm]{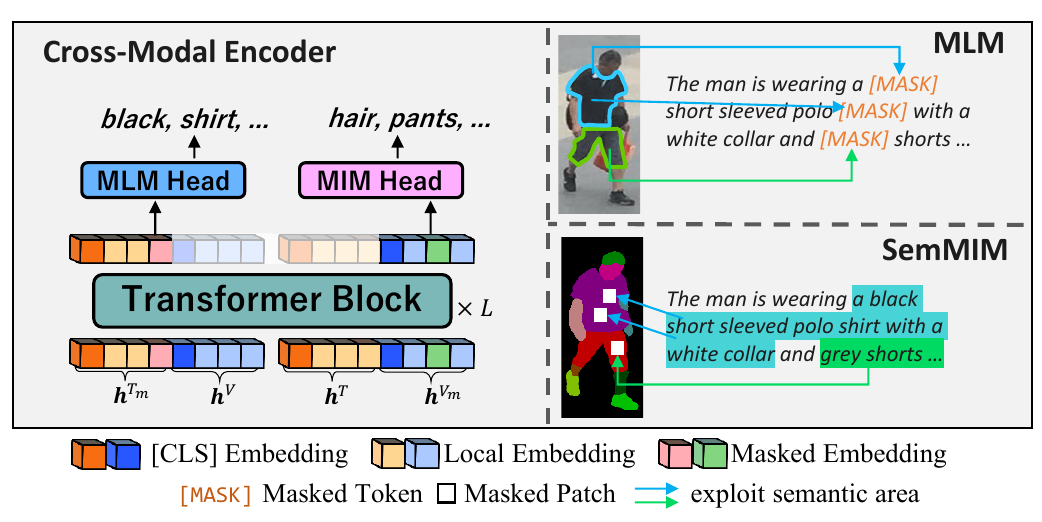}
   \caption{(Left) Cross-Model Encoder of BiLMa. (Right) MLM and our SemMIM. BiLMa enables a network to exploit visual/textual semantic area corresponding to masked textual/visual tokens via MLM/SemMIM.}
\label{fig:mlmmim}
\end{figure}

BiLMa framework is illustrated in Figure \ref{fig:model} and \ref{fig:mlmmim}.
When we train TBPReID models with this framework, not only text tokens but also image tokens are randomly masked, then their labels are predicted by unmasked image and text tokens.
More specifically, unmasked image and text tokens are fed into Cross-Modal Encoder (CME) to produce vectors for predicting the labels of masked image and text tokens.
The model parameters are optimized via jointly minimizing MLM loss ({\it cf}., \S 3.1) and MIM loss detailed later.

\noindent\textbf{Cross-Modal Encoder (CME).} 
As shown in the left of Figure \ref{fig:mlmmim}, CME consists of $L$-layer Transformer blocks, MLM head, and Masked Image Modeling (MIM) head.
Given image/text encoder outputs $\boldsymbol{h}^{V/T}$ ({\it cf}., \S 3.1), we randomly mask a portion of them to obtain masked image/text embeddings $\boldsymbol{h}^{V_m/T_m}$.
Unmasked image tokens $\boldsymbol{h}^{V}$ and masked text tokens $\boldsymbol{h}^{T_m}$ are concatenated to be fed into the Transformer blocks, then the resulting tokens corresponding to the masked tokens are further fed into the MLM head to produce logit vectors for classification of masked words.
Similarly, unmasked text tokens $\boldsymbol{h}^{T}$ and masked image tokens $\boldsymbol{h}^{V_m}$ are concatenated to be fed into the same Transformer blocks, then the resulting tokens corresponding to the masked tokens are further fed into the MIM head to produce logit vectors for classification of masked image tokens.
Notice that we compose both MLM and MIM heads as multi-layer perceptrons of $2$-layer with GELU and layer normalization.
CME is removed during inference stage.

\subsection{Semantic Masked Image Modeling (SemMIM)}
To make text-to-image local-matching more tractable, we further propose a novel MIM method, named Semantic MIM (SemMIM). 
In a nutshell, given the outputs of MIM head ({\it i.e.}, logit vectors corresponding to the masked image tokens) and their ground truth semantic labels ({\it e.g.}, hair, pants as shown in the right of Figure \ref{fig:mlmmim}), SemMIM optimize a model so as to minimize the loss of token label classification.
Formally, the MIM loss $\mathcal{L}_{mim}$ is the sum of Cross-Entropy between masked image tokens and its semantic labels, which is defined as follows:
\vspace{-2mm}
\begin{equation}
\small
    \mathcal{L}_{mim}=-\frac{1}{|\mathcal{M}_v||\mathcal{C}|}\sum_{i\in \mathcal{M}_V}\sum_{j\in |\mathcal{C}|}y_j^i\log\frac{\exp(m_{i,j}^{V_m})}{\sum_{k=1}^{|\mathcal{C}|}\exp(m_{i,k}^{V_m})},
    \vspace{-1mm}
\end{equation}
where $\mathcal{M}_v$ denotes the set of masked image tokens and $\mathcal{C}$ is the label set for tokens.
$y_j^i$ is 1 if the true label of $i$-th masked image token is $j$-th label in $\mathcal{C}$, and 0 otherwise.
$\{m_{i,k}^{V_m}\}_{j=1}^{|\mathcal{C}|}$ is the probability of $j$-th label in $\mathcal{C}$ of $i$-th masked image token. 
\par
A straightforward approach to obtain such semantic labels is manual annotation, which is apparently costly and even error-prone.
Therefore, we propose to introduce SOTA human parsing models to automatically give semantic labels to tokens.
Specifically, given an human parser $\phi$, we feed all the training images to $\phi$ to obtain pixel-wise semantic labels.
For each token that corresponds to an image token, its semantic label is determined as the most frequent label within the token.
In this work we employ a SOTA human parser \cite{li2020self} as $\phi$.
Exemplar parsing results are shown in \S A.4 of our supplemental material .

This method enables to exploit the textual semantic area corresponding to masked image tokens, and make ties between them stronger.
This exploitation process is illustrated in the bottom-right of Figure \ref{fig:mlmmim}.
Multi-task learning of MLM and SemMIM can achieve BiLMa, both image-to-text and text-to-image local-matching.

\subsection{Loss Function}
We train our model via minimizing the following loss $\mathcal{L}$:
\vspace{-2mm}
\begin{equation}
    \mathcal{L} = \mathcal{L}_{id} + \mathcal{L}_{sdm} + \alpha\mathcal{L}_{mlm} + \beta\mathcal{L}_{mim}.
    \vspace{-1mm}
\end{equation}
$\alpha$ and $\beta$ are hyperparameters to control the contribution of MLM and SemMIM, respectively. 

\begin{table*}[h]
  \begin{center}
{\footnotesize{
    \begin{tabular}{l|cccc|cccc|cccc}
    \toprule
    \multirow{2}{*}{\textbf{Method}} & \multicolumn{4}{|c|}{\textbf{CUHK-PEDES}} & \multicolumn{4}{c|}{\textbf{ICFG-PEDES}} & \multicolumn{4}{c}{\textbf{RSTPReid}}\\
     & R@1 & R@5 & R@10 & mAP & R@1 & R@5 & R@10 & mAP & R@1 & R@5 & R@10 & mAP\\
    \midrule
    ISANet~\cite{Yan2022ImageSpecificIS} & 63.92 & 82.15 & 87.69 & - & 57.73 & 75.42 & 81.72 & - & - & - & - & -\\
    LBUL~\cite{LBUL} & 64.04 & 82.66 & 87.22 & - & - & - & - & - & 45.55 & 68.20 & 77.85 & - \\
    AXM-Net~\cite{Farooq2021AXMNetIC} & 64.44 & 80.52 & 86.77 & 58.73 & - & - & - & - & - & - & - & - \\
    LGUR~\cite{Shao2022LearningGR} & 65.25 & 83.12 & 89.00 & - & 59.20 & 75.32 & 81.56 & - & - & - & - & - \\
    IVT~\cite{Shu2022SeeFS} & 65.59 & 83.11 & 89.21 & - & 56.04 & 73.60 & 80.22 & - & 46.70 & 70.00 & 78.80 & -\\
    CFine~\cite{Yan2022CLIPDrivenFT} & 69.57 & 85.93 & 91.15 & - & 60.83 & 76.55 & 82.42 & - & 50.55 & 72.50 & 81.60 & -\\
    IRRA~\cite{Jiang_2023_CVPR} & 73.38 & \textbf{89.93} & \textbf{93.71} & 66.13 & 63.46 & \textbf{80.25} & \textbf{85.82} & 38.06 & 60.20 & 81.30 & 88.20 & 47.17\\
    \midrule
    \textbf{BiLMa w/ SemMIM (Ours)} & \textbf{74.03} & 89.59 & 93.62 & \textbf{66.57} & \textbf{63.83} & 80.15 & 85.74 & \textbf{38.26} & \textbf{61.20} & \textbf{81.50} & \textbf{88.80} & \textbf{48.51}\\
    \bottomrule
    \end{tabular}
    }}
\end{center}
\vspace{-2mm}
\caption{Performance comparisons with state-of-the-art methods on CUHK-PEDES, ICFG-PEDES and RSTPReid datasets.}
\label{tab:sota}
\end{table*}

\begin{table*}[h]
\vspace{-1mm}
  \begin{center}
{\footnotesize{
    \begin{tabular}{cc|cccc|cccc|cccc}
    \toprule
    \multicolumn{2}{c|}{\textbf{Components}} & \multicolumn{4}{c|}{\textbf{CUHK-PEDES}} & \multicolumn{4}{c|}{\textbf{ICFG-PEDES}} & \multicolumn{4}{c}{\textbf{RSTPReid}}\\
     MLM & SemMIM & R@1 & R@5 & R@10 & mAP & R@1 & R@5 & R@10 & mAP & R@1 & R@5 & R@10 & mAP\\
    \midrule
     & & 73.01 & 88.92 & 93.58 & 65.62 & 63.09 & 80.00 & 85.62 & 37.99 & 59.50 & 80.55 & 88.35 & 47.06 \\
     \ding{51} & & 73.16 & 89.52 & \textbf{93.63} & 66.00 & 63.60 & \textbf{80.29} & 85.70 & 38.12 & 59.05 & 80.35 & 87.95 & 46.29 \\
     & \ding{51} & 73.55 & 89.41 & 93.54 & 66.28 & 63.08 & 80.11 & 85.63 & 37.97 & 59.40 & 80.70 & 87.35 & 46.05 \\
     \ding{51} & \ding{51} & \textbf{74.03} & \textbf{89.59} & 93.62 & \textbf{66.57} & \textbf{63.83} & 80.15 & \textbf{85.74} & \textbf{38.26} & \textbf{61.20} & \textbf{81.50} & \textbf{88.80} & \textbf{48.51} \\
    \bottomrule
    \end{tabular}
    }}
\end{center}
\vspace{-2mm}
\caption{Ablation study on each component of BiLMa on CUHK-PEDES, ICFG-PEDES and RSTPReid datasets.}
\label{tab:component}
\vspace{-3mm}
\end{table*}

\section{Experiment}
We conduct experiments on three popular benchmarks: CUHK-PEDES~\cite{CUHK-PEDES}, ICFG-PEDES~\cite{ICFG-PEDES}, and RSTP-Reid~\cite{RSTPReid}.
We employ widely-used Rank@$K$ ($K=1,5,10$, R@$K$ for brevity) and mean Average Precision (mAP) as evaluation metrics, in both of which the higher is better.
We compare our approach with 6 SOTA methods including ISANet~\cite{Yan2022ImageSpecificIS}, LBUL~\cite{LBUL}, AXM-Net~\cite{Farooq2021AXMNetIC}, LGUR~\cite{Shao2022LearningGR}, IVT~\cite{Shu2022SeeFS}, CFine~\cite{Yan2022CLIPDrivenFT}, and IRRA~\cite{Jiang_2023_CVPR}.
Due to page limitations, we leave the details of benchmarks and our implementations (including the selection of the human parser) in our supplementary material.
\subsection{Comparisons with SOTA Models}
The overall results on each dataset are shown in Table \ref{tab:sota}.
For each dataset, we tuned the patch mask rate $m_p$ and SemMIM loss weight $\beta$ using a grid search and report the best results, while other results are described in \S A.6.
Following~\cite{Jiang_2023_CVPR}, the token mask rate $m_t$ and MLM loss weight $\alpha$ is set to $m_t=0.15$ and $\alpha=1.0$.

We can clearly see that our approach (BiLMa w/ SemMIM) achieves the best Rank@1 and mAP on all the datasets.
Particularly, compared to the best scores of existing methods, Rank@1 of our approach on CUHK-PEDES is 0.56\% higher while mAP of ours on ICFG-PEDES is 0.37\% better.
On RSTPReid, ours achieves SOTA for all the metrics including Rank@1,5,10 and mAP.
These results indicate the superiority and the generalization ability of our proposed approach.


\subsection{Ablation Study}
Next, we analyze the contribution of our proposals.
Table \ref{tab:component} shows the results of our ablative models on three datasets.
From this table, we can observe that using both MLM and SemMIM ({\it i.e.}, BiLMa framework) tend to achieve the best performance, indicating the good compatibility of SemMIM with MLM.
Notice that our SemMIM can be solely used without MLM.
Interestingly, in several cases, our model with only SemMIM outperforms the model with only MLM, which implies the strong ability of SemMIM for TBPReID model training.
We also observe that our SemMIM outperforms other three MIM methods, detailed in \S A.5 of the supplemental material due to page limitations. 

\subsection{Qualitative Analysis}
\begin{figure}[htbp]
\vspace{-3mm}
\centering
\includegraphics[width=8cm]{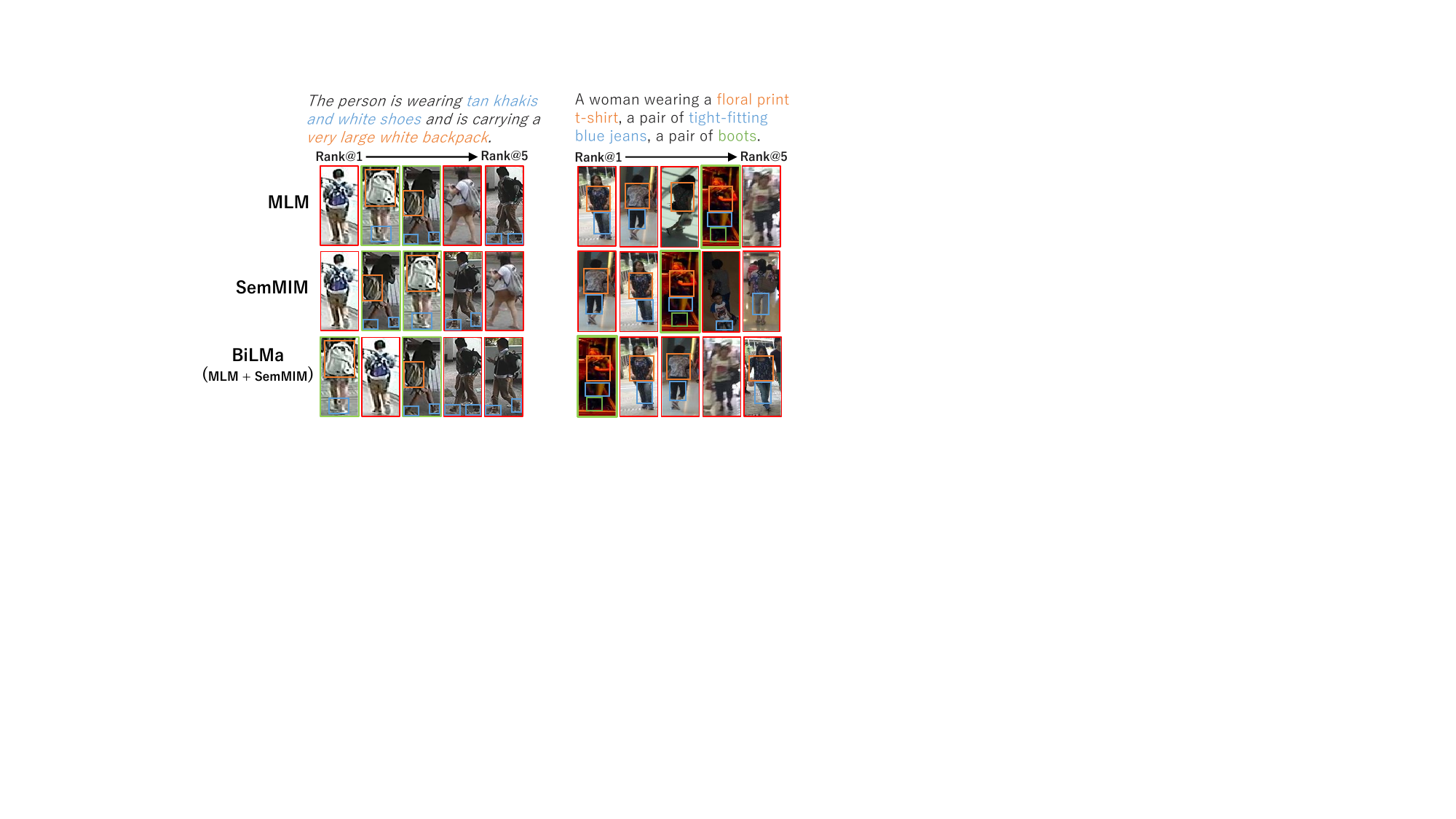}
   \caption{Comparison of top-5 retrieved results on CUHK-PEDES between ablative models with only MLM or SemMIM and BiLMa with both MLM and SemMIM for each text query.}
\label{fig:quantative}
\vspace{-6mm}
\end{figure}

Figure \ref{fig:quantative} shows two top-5 retrieval results of our model (3rd row) given a textual query shown at the top.
Results of 1st and 2nd rows are our ablative models comprising only MLM or SemMIM, respectively.
An image with a green frame is true positive while the one with a red frame is false positive.
For clarity, phrases and their corresponding tokens are made the same color.
These results show that our BiLMa w/ SemMIM can retrieve correct person more correctly.
One possible reason of this superiority is that BiLMa can discriminate \texttt{very large white backpack}, \texttt{boots}, \texttt{white backpack}, \texttt{tight-fitting}, and \texttt{boots} correctly.

\section{Conclusion}
In this work, we proposed Bidirectional Local-Matching (BiLMa) framework that jointly optimizes MLM and MIM in TBPReID model training.
We also proposed Semantic Masked Image Modeling (SemMIM) to make text-to-image local-matching more tractable.
Experiments on three TBPReID benchmarks demonstrate that our BiLMa w/ SemMIM achieves SOTA Rank@1 and mAP on all the datasets.
As our future research, we plan to (1) find more helpful Masked Image/Language Modeling strategies, (2) investigate the influence of human parser's errors and consider a way to cover them.

\newpage
{\footnotesize
\bibliographystyle{ieee_fullname}
\bibliography{main_revised}
}

\appendix
\section{Appendix}
\begin{table*}[h]
  \begin{center}
    {\footnotesize{
\begin{tabular}{l|ccc|ccc|ccc}
\toprule
\multirow{2}{*}{\textbf{Datasets}} & \multicolumn{3}{c|}{\textbf{IDs}} & \multicolumn{3}{c|}{\textbf{Images}} & \multicolumn{3}{c}{\textbf{Textual Descriptions}}\\
& train & test & val & train & test & val & train & test & val\\
\midrule
CUHK-PEDES & 11003 & 1000 & 1000 & 34054 & 3074 & 3078 & 68126 & 6156 & 6158\\
ICFG-PEDES & 3102 & 1000 & 0 & 34674 & 19848 & 0 & 34674 & 19848 & 0\\
RSTPReid & 3701 & 200 & 200 & 18505 & 1000 & 1000 & 37010 & 2000 & 2000\\
\bottomrule
\end{tabular}
}}
\end{center}
\caption{Dataset statistics of CUHK-PEDES, ICFG-PEDES and RSTPReid.}
\label{tab:statics}
\end{table*}
\subsection{Datasets Details} \label{sec:dataset}
In this section, we introduce three benchmark datasets in Text-based Person Re-identification (TBPReID).
The dataset statistics are shown in Table \ref{tab:statics}.

\noindent\textbf{CUHK-PEDES~\cite{CUHK-PEDES}.} This is the first introduced dataset for TBPReID, which contains 40206 images and 80412 textual descriptions for 13003 IDs.

\noindent\textbf{ICFG-PEDES~\cite{ICFG-PEDES}.} The second introduced dataset for TBPReID, which contains 54522 images for 4102 IDs. Each image has only one corresponding
textual description.

\noindent\textbf{RSTPReid~\cite{RSTPReid}.} The newly introduced dataset for TBPReID, which contains 20505 images for 4101 IDs from 15 cameras.
Each ID has five corresponding images taken by different cameras and each image has two corresponding textual descriptions.

\subsection{Loss Equation} \label{sec:loss}
In this section, we introduce equations of SDM loss and ID loss used in IRRA.
Please refer to the original papers \cite{Jiang_2023_CVPR,Zheng2017DualpathCI} for more details.

\noindent\textbf{SDM Loss.}
SDM losss is represented as follows:
\begin{equation}
    p_{i,j}=\frac{\exp(\text{sim}(\boldsymbol{h}_{cls,i}^V,\boldsymbol{h}_{sos,j}^T)/\tau)}{\sum_{k=1}^N \exp(\text{sim}(\boldsymbol{h}_{cls,i}^V,\boldsymbol{h}_{sos,k}^T)/\tau)},
    \label{eq:sdm_1}
\end{equation}
\begin{equation}
    \mathcal{L}_{i2t}=\text{KL}(\boldsymbol{p}_i||\boldsymbol{q}_i)=\frac{1}{N}\sum_{i=1}^N\sum_{j=1}^N p_{i,j}\log\left(\frac{p_{i,j}}{q_{i,j}+\epsilon}\right),
    \label{eq:sdm_2}
\end{equation}
\begin{equation}
    \mathcal{L}_{sdm}=\mathcal{L}_{i2t}+\mathcal{L}_{t2i},
    \label{eq:sdm_3}
\end{equation}
where $N$ is mini-batch size, $\tau$ is a temperature hyperparameter which controls the probability distribution peaks.

\noindent\textbf{ID Loss.}
ID loss is represented as follows:
\begin{equation}
\begin{split}
    \mathcal{L}_{id}=-(\boldsymbol{y}_{id}\log(\text{softmax}(\boldsymbol{W}_{id}\boldsymbol{v}_{cls}))\\
    +\boldsymbol{y}_{id}\log(\text{softmax}(\boldsymbol{W}_{id}\boldsymbol{t}_{cls}))),
\end{split}
\label{eq:idloss}
\end{equation}
Where $\boldsymbol{W}_{id}$ is a shared transformation matrix to classify the different persons and $\boldsymbol{y}_{id}$ is the ground true identity.

\subsection{Implementation Details} \label{sec:impl}
We conduct our experiments on a single NVIDIA A100 80GB GPU.
For an image and text encoder, we use pretrained CLIP-ViT-B/16 and CLIP text encoder respectively.
Table \ref{tab:hypara} lists hyperparameters for our experiments. 
In training, we adopt three image data augmentation methods of random horizontally flipping, random crop with padding and random erasing.
For annotating all the training images with semantic labels, we use three kinds of human parsers\footnote{Human parsers are publicly available in \url{yanhttps://github.com/GoGoDuck912/Self-Correction-Human-Parsing}} trained on ATR~\cite{7053923}, LIP~\cite{8100198} or PPP~\cite{10.1109/CVPR.2014.254} introduced in \cite{li2020self}.
The number of the semantic classes is 18, 20, 7 in ATR, LIP and PPP, respectively.

\begin{table}[htbp]
  \begin{center}
    {\footnotesize{
\begin{tabular}{lc}
\toprule
\textbf{Hyperparameter} & \textbf{Value} \\
\midrule
vocab size of tokenizer & 49408 \\
hidden size & 512\\
attention heads of CME & 8\\
Transformer blocks in CME, $L$ & 4\\
input image size & $384\times128$\\
textual token sequence & 77 \\
batch size & 32 \\
epoch & 60 \\
learning rate & $1\times 10^{-5}$ \\
Adam $\alpha$ & 0.9 \\
Adam $\beta$ & 0.999 \\
learning rate decay & cosine \\
warm-up & $1\times10^{-6}\rightarrow1\times10^{-5}$ \\
 & linearly at first 5 epochs \\
tempreture in SDM loss & 0.02 \\
token mask rate & 0.15 \\
MLM loss weight $\alpha$ & 1.0 \\
\bottomrule
\end{tabular}
}}
\end{center}
\caption{Hyperparamters in our experiments.}
\label{tab:hypara}
\end{table}

\subsection{Exemplar Parsing Results}
\begin{figure}[htbp]
\centering
\includegraphics[width=7.5cm]{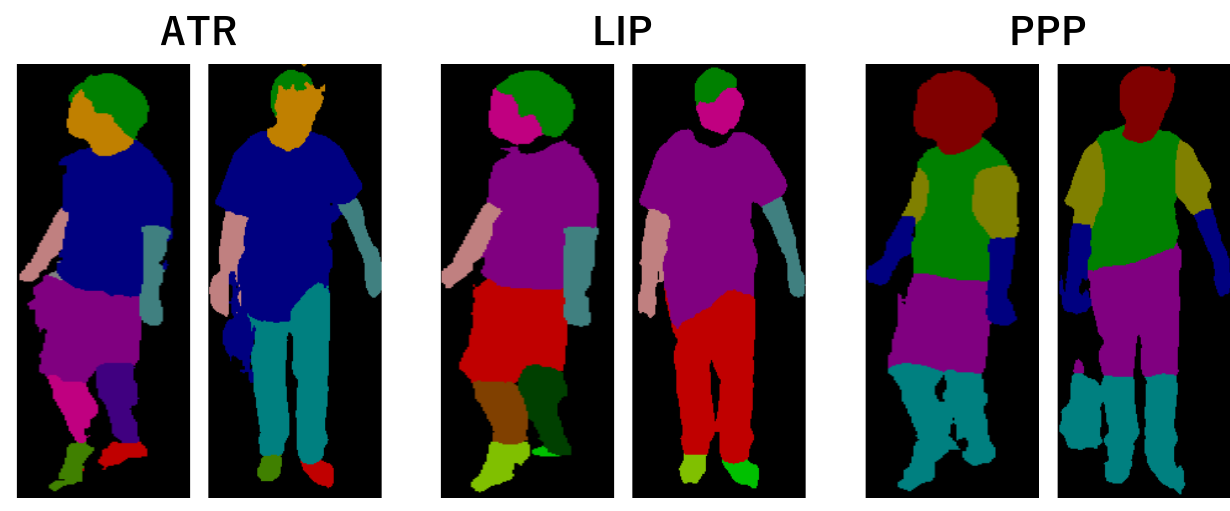}
   \caption{Labelling results of each human parser trained on ATR, LIP and PPP in CUHK-PEDES.}
\label{fig:hp}
\end{figure}

Labelling results of each human parser in CUHK-PEDES are shown in Fig.\ref{fig:hp}.
We observe that all human-parser can label some CUHK-PEDES samples with high-quality.

\subsection{Existing MIM Strategies}
A straight forward approach to perform text-to-image local-matching is to adapt existing Masked Image Modeling (MIM)~\cite{9880205,Bao2021BEiTBP,Cao2022HowTU,Wang_2023_CVPR}, which are formulated as a patch reconstruction problem.
Therefore, for text-to-image local-matching in BiLMa, we attempt three simple MIM methods in similar ways to existing methods:

\noindent\textbf{Pixel-level MIM} is to reconstruct original RGB values of masked image tokens. In the case of patch size $P\times P$, a model predict $P^2$ values per masked image token.

\noindent\textbf{Patch-level MIM} is to reconstruct RGB values avaraged within tokens of masked image tokens. A model predict one value per masked image token.

\noindent\textbf{Feature-level MIM} is to reconstruct embeddings of masked tokens.
In the case of $d$-dimentional embeddings, a model generate $d$ values vector per masked image token.

Mask rate and MIM loss weigh is set to $m=0.15$ and $\beta=1.0$. For training objectives, we use MSE loss, that is widely used in image reconstruction tasks, in Pixel- and Patch-level MIM, and KL-divergence in Feature-level MIM.

Table\ref{tab:res1} shows Rank@1 and mAP scores of the model with each MIM methods in CUHK-PEDES. 
Pixel-level and Patch-level MIM cannot obtain higher performances than IRRA and SemMIM.
Feature-level MIM outperforms IRRA slightly, but less than SemMIM.
These results show it is difficult to solve existing MIM methods by unmasked text embeddings.
\begin{table}[h]
  \begin{center}
    {\small{
\begin{tabular}{lcc}
\toprule
\textbf{MIM Method} & Rank@1 & mAP\\
\midrule
w/o MIM & 73.38 & 66.13\\
\midrule
Pixel-level & 72.86 & 65.61\\
Patch-level & 73.07 & 66.01\\
Feature-level & 73.52 & 66.20\\
\midrule
SemMIM (Ours) & \textbf{74.03} & \textbf{66.57}\\
\bottomrule
\end{tabular}
}}
\end{center}
\caption{Performance comparison with three existing MIM methods in CUHK-PEDES.}
\label{tab:res1}
\end{table}

\subsection{Additional Ablations} \label{sec:add}
We search hyperparameters, mask rate $m$ and SemMIM loss weight $\beta$, with high performance using grid search.
In our main paper, we reported only the best set, while other results are reported in this section.

\subsubsection{Mask Rate}
\begin{figure}[htbp]
\centering
\includegraphics[width=8cm]{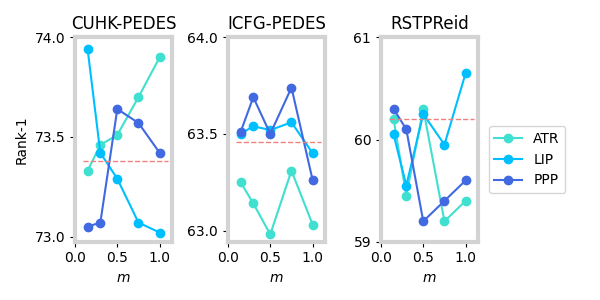}
   \caption{Rank@1 transition of each human parser when changing mask rate $m$.}
\label{fig:m}
\end{figure}

We conduct our experiment in the range of mask rate $m=\{0.15,0.30,0.50,0.75,1.0\}$ with the SemMIM loss weight $\beta=1.0$ using three human parsers on three benchmarks.
Figure \ref{fig:m} shows the Rank@1 transitions of each human parser.
The red lines are Rank@1 accuracies of baselines ({\it i.e.} IRRA).

\subsubsection{SemMIM Loss Weight}
\begin{figure}[htbp]
\centering
\includegraphics[width=8cm]{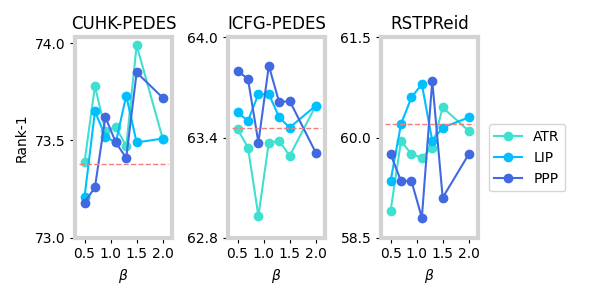}
   \caption{Rank@1 transition of each human parser when changing SemMIM loss weight $\beta$ in mask rate $m=0.15$.}
\label{fig:m015}
\end{figure}

\begin{figure}[htbp]
\centering
\includegraphics[width=8cm]{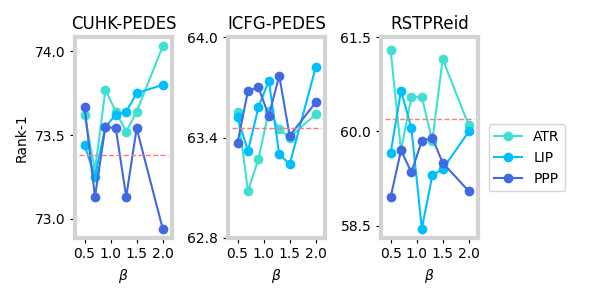}
   \caption{Rank@1 transition of each human parser when changing SemMIM loss weight $\beta$ in mask rate $m=0.50$.}
\label{fig:m050}
\end{figure}

We conduct our experiment in the range of SemMIM loss weight $w=\{0.5,0.7,0.9,1.1,1.3,1.5,2.0\}$ with the mask rate $m=0.15,0.5$ using three human parsers on three benchmarks.
Figure \ref{fig:m015} shows the Rank@1 transitions of each human parser in mask rate $m=0.15$. The red lines are Rank@1 accuracies of baseline ({\it i.e.} IRRA).
Similarly, Figure \ref{fig:m050} shows the Rank@1 transitions of each human parser in mask rate $m=0.50$. The red lines are Rank@1 accuracies of baseline ({\it i.e.} IRRA).

\end{document}